\pdfoutput=1

\documentclass[11pt]{article}

\usepackage[]{acl}

\usepackage{times}
\usepackage{latexsym}

\usepackage[T1]{fontenc}

\usepackage[utf8]{inputenc}

\usepackage{microtype}

\usepackage{inconsolata}

\usepackage{amsmath,amssymb,amsfonts,amsthm}
\usepackage{mathtools}
\usepackage{url,enumitem}
\usepackage{booktabs}
\usepackage{xspace}
\usepackage{graphicx}
\usepackage{subcaption}
\usepackage{comment}
\usepackage[normalem]{ulem}
\usepackage{mathtools}
\usepackage{bm}
\usepackage{svg}
\usepackage{algorithm,algorithmicx,algpseudocode}
\usepackage{multicol}
\usepackage{multirow}
\usepackage{listings}
\usepackage{longtable}
\usepackage{tabularx}
\usepackage{pifont}
\usepackage{bbding}
\newcommand{\cmark}{\ding{51}}
\newcommand{\xmark}{\ding{55}}
\usepackage{xcolor}
\definecolor{right}{HTML}{95061E}
\definecolor{left}{HTML}{014E81}
\usepackage{stfloats}
\usepackage{balance}
\usepackage{flushend}

\usepackage{dsfont}

\newcommand{\Sref}[1]{\S\ref{#1}}

\newcommand{\ourmethod}[1]{$\textsc{P}^3\textsc{Sum}$}

\title{
\ourmethod{}: Preserving Author's Perspective in News Summarization \\with Diffusion Language Models}

\author{
Yuhan Liu$^{*\clubsuit}$ \ \ \ \ \ \ \ Shangbin Feng$^{*\spadesuit}$ \ \ \ \ \ \ \ Xiaochuang Han$^{\spadesuit}$ \\ \bf Vidhisha Balachandran$^{\heartsuit}$  \ \ \
Chan Young Park$^{\heartsuit}$ \ \ \ Sachin Kumar$^{\diamondsuit}$ \ \ \ Yulia Tsvetkov$^{\spadesuit}$\\
$^{\clubsuit}$Xi'an Jiaotong University,
$^{\spadesuit}$University of Washington\\ 
$^{\heartsuit}$Carnegie Mellon University,
$^{\diamondsuit}$Allen Institute for AI \\    
\href{mailto:lyh6560@stu.xjtu.edu.cn}{\texttt{lyh6560@stu.xjtu.edu.cn}};\ \href{mailto:shangbin@cs.washington.edu}{\texttt{shangbin@cs.washington.edu}}
}

\begin{document}
\maketitle
\def\thefootnote{*}\footnotetext{These authors contributed equally to this work.}\def\thefootnote{\arabic{footnote}}
\begin{abstract}

In this work, we take a first step towards designing summarization systems that are faithful to the author's intent, not only the semantic content of the article. 
Focusing on a case study of \emph{preserving political perspectives in news summarization}, we find that existing approaches alter the political opinions and stances of news articles in more than 50\% of summaries, misrepresenting the intent and perspectives of the news authors. We thus propose \ourmethod{}, a diffusion model-based summarization approach controlled by political perspective classifiers.
In \ourmethod{}, the political leaning of a generated summary is iteratively evaluated at each decoding step, and any drift from the article's original stance incurs a loss back-propagated to the embedding layers, steering the political stance of the summary at inference time.
Extensive experiments on three news summarization datasets demonstrate that \ourmethod{} outperforms state-of-the-art summarization systems and large language models by up to 13.7\% in terms of the success rate of stance preservation, with competitive performance on standard metrics of summarization quality. Our findings present a first analysis of preservation of pragmatic features in summarization, highlight the lacunae in existing summarization models---that even state-of-the-art models often struggle to preserve author's intents---and develop new summarization systems that are more faithful to author's perspectives.\footnote{Code and data are publicly available at \href{https://github.com/lyh6560new/P3Sum}{https://github.com/lyh6560new/P3Sum}.}

\end{abstract}

\section{Introduction}


What constitutes a faithful summary? 
In addition to preserving factual consistency---the focus of much prior work \citep{kryscinski-etal-2020-evaluating,goyal-durrett-2020-evaluating,wang-etal-2020-asking,pagnoni-etal-2021-understanding, feng2023factkb,tam-etal-2023-evaluating}---a good summarization system should preserve the \emph{writer's voice}---the style, intent, and points of view conveyed by the authors. However, such subtle pragmatic cues are harder to extract and control for by existing models \citep{borji2023categorical}, and it remains underexplored whether existing summarization systems generate summaries that are \emph{faithful} to the opinions and perspectives of the authors. Moreover, though language models (LMs) have been widely applied to many summarization tasks,
they inevitably contain political biases and such biases could further impact downstream tasks \citep{feng-etal-2023-pretraining}.  So we hypothesize that summarization systems built on top of LLMs would propagate biases further, but not necessarily align them with stances in the source text. Specifically in the task of summarization, instead of ``de-biasing'' and generating only neutral summaries, we argue that a good summarization system should \emph{preserve the perspectives} of the authors in generated news summaries. 

\begin{figure}[t]
	\centering

        \includegraphics[width=1.0\linewidth]{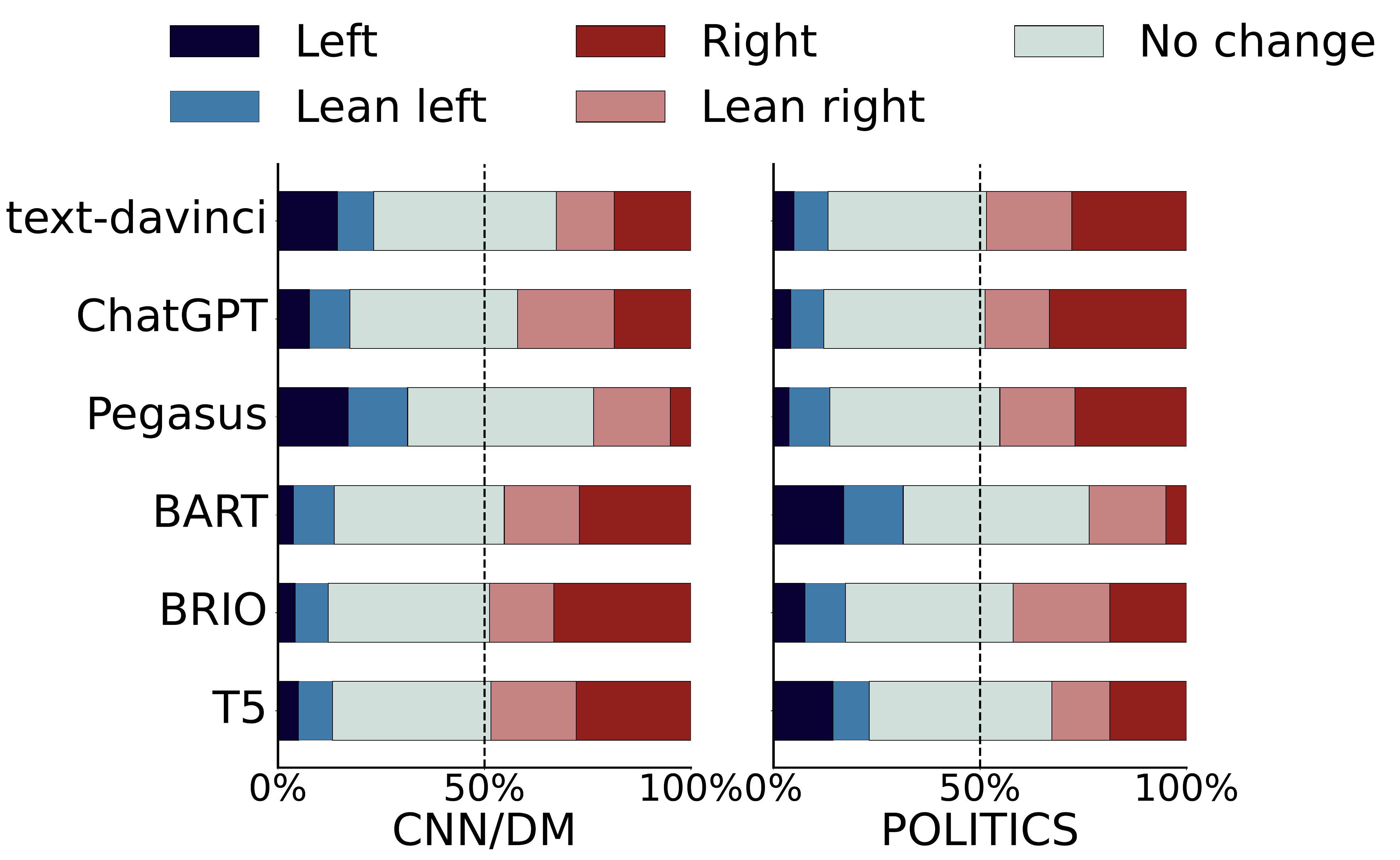}
	\caption{
        Changes in political stances between the summary and the article. The political perspective classifier produces \emph{left}, \emph{center}, or \emph{right} labels for each text sequence. 
        Left (or Right) indicates a shift in summary stance towards left (or right) by 2 units while Lean Left (Or Lean Right) indicates a shift by 1 unit. No change indicates that there is no difference in the political leaning of the summary and the context. \textbf{Our study shows that existing approaches alter the stances of news articles in more than 50\% of cases across both datasets}.
	}
	\label{teaser}
        \vspace{-20pt}
\end{figure}

To this end, we first evaluate to what extent summarization systems and LLMs preserve political stances in generated summaries, by employing a state-of-the-art political perspective evaluator \citep{liu-etal-2022-politics} to quantify the gap between stances in news articles and summaries. (\Sref{sec:analysis}) We identify that existing summarization systems and LLMs \emph{do} alter opinions and perspectives in the original document, resulting in shifting stances in more than 50\% of summaries, with around 25\% drifting to the partisan extremes (Figure \ref{teaser}). This highlights a new, underexplored concern with current LLMs as they fail to preserve the intents and perspectives of the authors of news documents during summarization, potentially misinforming the readers.

To address this issue, we propose \ourmethod{}, a summarization model aiming to \textbf{P}reserve the \textbf{P}olitical \textbf{P}erspectives of news articles. (\Sref{sec:method}) \ourmethod{} employs a non-autoregressive diffusion language model with modular control capabilities to steer the generated summary towards the same perspective of the news article. Specifically, we first fine-tune a diffusion language model \citep{mahabadi2023tess,han2023ssd,han-etal-2023-ssd} on summarization data.
During inference, the generated summary is evaluated by a political stance classifier \citep{liu-etal-2022-politics} at each step, compared to the target stance in the source document while summary generation is steered towards the target stance. 
Our primary motivation to use diffusion models is that they allow us to (1) apply the stance classifier on the whole summary at each decoding step, rather than on a prefix generated autoregressively \citep{kumar-etal-2022-gradient}, and (2) seamlessly incorporate various pretrained classifiers without adaptation, to carefully steer generation process.   
Thus, 
as an inference-time approach based on diffusion models and controllable text generation \citep{kumar2021controlled,li2022diffusion,han-etal-2023-ssd,han2023ssd,mahabadi2023tess,austin2021structured,strudel2022self,dieleman2022continuous}, \ourmethod{} alleviates the need for additional training or pretraining, handles news articles from different ideological stances, and is compatible with future classifiers of author perspectives.


Extensive experiments on three news datasets demonstrate that \ourmethod{} greatly outperforms baselines in preserving the political stances of news articles while maintaining good summarization utility. Specifically, \ourmethod{} is at least 13.7\%, 2.9\%, and 1.6\% better in perspective preservation on CNN/DM \citep{nallapati2016abstractive}, XSUM \citep{narayan2018don}, and POLITICS \citep{liu-etal-2022-politics}, outperforming popular summarization systems \citep{raffel2020exploring,liu-etal-2022-brio,zhang2020pegasus} and large language models \citep{touvron2023llama,penedo2023refinedweb,vicuna2023}. In addition, \ourmethod{} obtains ROUGE scores and abstractiveness metrics that are only slightly lower than state-of-the-art systems, while qualitative analysis highlights \ourmethod{}'s effectiveness in generating high-quality, perspective-preserving summaries. We envision \ourmethod{} as a first step towards summarization systems that are faithful to the intents and perspectives of the authors.



\begin{table}[t]
    \centering
    \resizebox{0.5\linewidth}{!}{
    \begin{tabular}{@{}lcc@{}}
        \toprule[1.5pt]
        \textbf{\textsc{change}}&  
        \textbf{\textsc{cnn/dm}}& 
        \textbf{\textsc{xsum}}\\
        \midrule[0.75pt]
         Left&
         20.6&
         5.0\\
         Lean left&
         13.2&
         3.8\\
         No change&
         43.0&
         39.2\\
         Lean right&
         15.8&
         14.2\\
         Right&
         7.4&
         37.8\\
         \bottomrule[1.5pt]
    \end{tabular}
    }
    \caption{Changes (\%) in political stances between the gold summary annotations and the news article. Around 57\% to 60.8\% of reference summaries in news summarization datasets alter author perspectives.}
    \label{tab:gold}
    \vspace{-15pt}
\end{table}

\section{Examining Perspective Preservation}
\label{sec:analysis}
Given a news article, the generated summary should preserve the authors' political perspectives in the document. However, existing models are not designed to control for author intent or perspectives, and we first investigate to which extent summarization systems and large language models alter the perspectives in the generated summaries.

To this end, we measure the political leaning of the generated summaries and compare them to the political stances of original articles, using 500 randomly chosen news articles from the \textsc{CNN/DM} \citep{nallapati2016abstractive} and \textsc{politics} \citep{liu-etal-2022-politics} datasets\footnote{All data are sampled from the test sets of the datasets}. 
We use a political perspective evaluator \citep{liu-etal-2022-politics} to quantify political stances of summaries and news articles (mapping text sequences to \emph{left}, \emph{center}, or \emph{right}), investigating the change in political leanings with six summarization models and LLMs: \textsc{GPT-3.5} (\textsc{text-davinci-003}), \textsc{chatgpt} (\textsc{gpt-3.5-turbo}), \textsc{pegasus} \citep{zhang2020pegasus}, \textsc{bart} \citep{lewis2020bart}, \textsc{brio} \citep{liu-etal-2022-brio}, and \textsc{t5} \citep{raffel2020exploring}. We then determine the perspective gap between the summary and the news article. 

As shown in Figure \ref{teaser}\footnote{For more specific numbers, please refer to Appendix \ref{changes_cont}}
, current summarization systems struggle to provide faithful summaries and significantly alter political perspectives. Concretely, the political stance of the generated summary is different from the news article in more than 50\% of cases across different models, while around 25\% drift to partisan extremes.

Besides, we also examine the political perspective of reference summaries provided in well-established summarization datasets, namely CNN/DM and XSUM in Table \ref{tab:gold}, and find that more than $50\%$ of them also alter the stances of the given article. Although these human-written or annotated summaries are considered gold standards for summarization tasks and are used for both training and evaluation, they hardly preserve the original political perspectives, incorporating another layer of data bias into the training and evaluation process.

As a result, how to develop summarization approaches that are faithful to the authors' perspectives in the news document remains an open research question.

\vspace{-5pt}

\section{\ourmethod{}}
\vspace{-5pt}
\label{sec:method}
We propose \ourmethod{}, a diffusion model that steers the political stance of the generation towards the news article at inference time with an off-the-shelf classifier. 
Given a news article $\boldsymbol{d}$, \ourmethod{} aims to generate a summary $\boldsymbol{s}$ that preserves the original political stance of the article. We first finetune a diffusion-based language model on summarization datasets. At decoding time, we employ a political stance classifier to steer the generated summary by incorporating the gradient from the classifier, ensuring that the political stance of the generation is consistent with the original article. 
\begin{figure*}[t]
	\centering

 \includegraphics[width=0.9\linewidth]{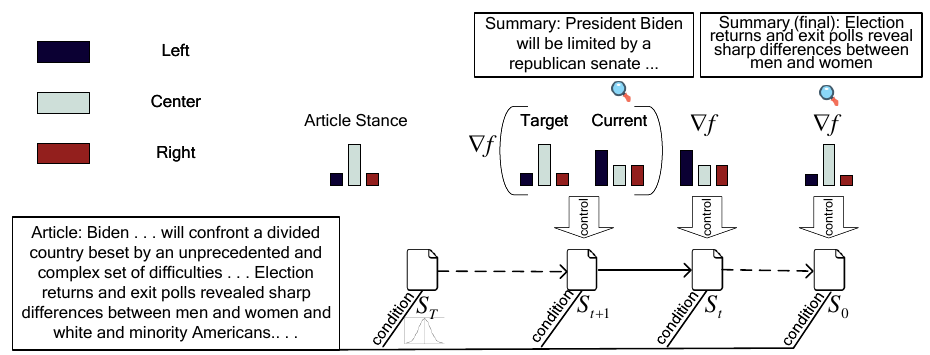}
	\caption{ During inference time, we iteratively refine the noisy logits and guide the perspective towards the original political stance by modular control. At each time step, we compare the stance between the current version of the summary and the given article. Then  a loss will be calculated if there is any inconsistency, and the corresponding gradients will be backpropagated to steer the generation for the following steps. At training time, we add progressive noise to $\mathbf{S}_0$ and learn to predict $\mathbf{S}_0$ from each noisy $\mathbf{S}_t$.}
	\label{fig:overview}
        \vspace{-15pt}
\end{figure*}
\subsection{Diffusion Model Finetuning}
At a high level, a diffusion model performs forward diffusion by adding noise to the original data and then learns to reconstruct the input \cite{sohl2015deep,ho2020denoising,chen2022analog,han-etal-2023-ssd,han2023ssd,mahabadi2023tess}. During inference time, we use the learned model to iteratively reconstruct from noisy representations and obtain high-quality generations.  To preserve the political stance, we modify the decoding process by incorporating the gradients from an external political classifier iteratively to guide the model generation. 

\paragraph{Continuous Data Representation}
Following \citet{han-etal-2023-ssd}, we define a function $\operatorname{logits-initialization}(\cdot)$ to obtain a logits representation over the model's vocabulary $\mathcal{V}$, mapping each discrete tokens of the news context and summary into continuous space. We map a token $w$ to $\Tilde{\boldsymbol{w}} \in \{ -K, +K \}^{|V|}$ as follows:
\vspace{-0.5em}
\begin{align}
\Tilde{w}^{(j)} = \begin{cases}
+K \text{ when } w = \mathcal{V}^{(j)}\\
-K \text{ when } w \neq \mathcal{V}^{(j)}                                  
\end{cases} 
\nonumber
\end{align}
where $V^{(j)}$ denotes the $j$-th token in the vocabulary and $K$ is a pre-defined scalar hyperparameter.


\paragraph{Forward Diffusion} For each passage $\boldsymbol{d}$ and gold summary $\boldsymbol{s}$, we concatenate them to form a sequence $\boldsymbol{w} = (w_1, \dots, w_L)$. 
We adopt non-autoregressive modeling \citep{mahabadi2023tess} which feeds the entire sequence into the model to better handle long article contexts. Let $\mathbf{S}_0 = (\Tilde{\boldsymbol{w}_1}, \dots, \Tilde{\boldsymbol{w}_L}) \in \{\pm K\}^{L \times |V|}$ be the logit representations of $\boldsymbol{w}$. Each step in the forward diffusion derives $\mathbf{S}_t$ by: 
$\mathbf{S}_t = \sqrt{\Bar{\alpha}_t} \mathbf{S}_0 + \sqrt{1-\Bar{\alpha}_t} \bm{\epsilon}_t$
where $t \in (1, T)$, $\boldsymbol{\epsilon}_t \sim \mathcal{N}(\boldsymbol{0}, K^2\mathbf{I})$, and $\Bar{\alpha}_t \to 0$ as $t \to T$ following a predefined schedule.
At step $T$, $\operatorname{sm}(\mathbf{S}_T)$ are fully noisy simplexes over $V$ (we use $\operatorname{sm}$ as a shorthand for softmax). 

\paragraph{Reverse Process}
Based on the noisy representation $\mathbf{S}_t$ (or noisy simplex  $\operatorname{sm}(\mathbf{S}_t)$) and a current timestep $t$, we learn to reverse the forward process by predicting the original representation $\mathbf{S}_0$ with our model $\operatorname{Transformer}_\theta$. The predicted outputs are the output logits from the Transformer model $\theta$, denoted as $\hat{\mathbf{S}}_\theta(\mathbf{S}_t,t)$. 
\begin{align}
    \hat{\mathbf{S}}_\theta(\mathbf{S}_t,t) =  \operatorname{Transformer}_\theta(\operatorname{sm}(\mathbf{S}_t), t)
    \label{func:predict}
\end{align}
We also apply self-conditioning \citep{chen2022analog} with a 50\% probability during prediction, re-computing $\mathbf{S}_t$ in Eq. \ref{func:predict} by:\footnote{See \citet{mahabadi2023tess} for more details.}
\vspace{-5pt}
\begin{align}
    \mathbf{S}_t = \frac12(\mathbf{S}_t+\hat{\mathbf{S}}_\theta(\mathbf{S}_{t},t))
    \nonumber
\end{align}
\paragraph{Loss Function}
After obtaining the model prediction $\hat{\mathbf{S}}_\theta(\mathbf{S}_t,t)$, we employ a cross-entropy loss between this predicted representation of $\mathbf{S}_0$ and the target summary tokens $\boldsymbol{w}$:
\vspace{-15pt}

\begin{align}
\mathcal{L}(\bm\theta) &= \mathbb{E}_{t, \mathbf{S}_0} \left[ -\sum_{i \in \mathbf{s}} \log p_{\bm{\theta}}(w_i | \mathbf{S}_t,t) \right]
\nonumber\\
&= \mathbb{E}_{t, \mathbf{S}_0} \left[ -\sum_{i \in \mathbf{s}} \log \operatorname{sm}[\hat{\mathbf{S}}_\theta(\mathbf{S}_t,t)]_{w_i} \right] \nonumber
\end{align}
\noindent where 
$\log p_{\bm{\theta}}(\cdot | \cdot)$ denotes the cross-entropy loss over the output logits of the transformer model $\theta$ that we are learning,\footnote{
For more details, see \citet{han-etal-2023-ssd,han2023ssd}.} and $i \in \mathbf{s}$ denotes whether this token belongs to summary $\mathbf{s}$.


\subsection{Perspective-Guided Decoding}
A diffusion language model generates the output sequence non-autoregressively by initializing a noise sequence  $\mathbf{S}_{T}$ and iteratively refining it through $\mathbf{S}_{t+1},\mathbf{S}_{t},\dots,\mathbf{S}_{0}$. 

Given an article as input, we initialize the summary as a noisy sequence $\mathbf{S}_T$ where each token is represented as a logit sampled from the normal distribution $\mathcal{N}(\boldsymbol{0}, K^2\mathbf{I})$. Using our learned model $\bm{\theta}$, we first obtain an estimated  output reconstructing from $\mathbf{S}_{T}$:
\vspace{-10pt}
\begin{align}
\mathbf{\hat{S}}_{\text{sc}, T} &= \hat{\mathbf{S}}_{\bm{\theta}}( \mathbf{S}_{T} , T),
\label{func:decode}
\end{align}

\paragraph{Self-Conditioning}
\citet{mahabadi2023tess} observe that self-conditioning \citep{chen2022analog} can improve the consistency between the model predictions and given context. Following their setting, for all steps $t<T$, we perform self-conditioning by mixing and leveraging the predictions from the previous time step in the current step. 
Let $\mathbf{S}_{t+1}$ denotes the incoming logits at $t$  from the previous time step $t+1$, and $ \hat{\mathbf{S}}_{sc,t+1}$ denotes the original estimation of the logits at time step $t+1$. We perform self-conditioning by computing the average of these representations and then pass to the model $\bm\theta$ for a prediction:  
\vspace{-10pt}
\begin{align}
 \mathbf{\hat{S}}_{\text{sc}, t} &= \hat{\mathbf{S}}_{\bm{\theta}}( \frac{  \mathbf{S}_{t+1} + \mathbf{\hat{S}}_{\text{sc}, t+1}  }{2} , t+1)
\nonumber
\end{align}
\paragraph{Modular Control}
We employ political bias classifiers to steer the generated summary toward the stances of the news article.
To guide \ourmethod{} to generate summaries with a target political leaning $y \in \{\textit{left}, \textit{center}, \textit{right}\}$, we use an external stance classifier $f_{\phi}(\cdot)$ that maps texts to the three stance labels and update our previous prediction $\mathbf{\hat{S}}_{\text{sc}, t}$ at each timestep $t$ guided by the gradients from the political stance classifier. 
\vspace{-5pt}
\begin{align}
   \mathbf{\hat{S}}_{\text{ctr}, t} = \mathbf{\hat{S}}_{\text{sc}, t} + \lambda \nabla_{\mathbf{\hat{S}}_{\text{sc}, t}} f_{\phi}(y \mid \operatorname{sm}(\mathbf{\hat{S}}_{\text{sc}, t})) 
    \label{equ:ctr}
\end{align}
where $\lambda$ is controlling learning rate, a hyperparameter governing the intensity of stance steering and the parameters of $\phi$ are frozen. This enables \ourmethod{} to iteratively steer the political stances of the generated summary toward the news article. \ourmethod{} employs a modular \emph{plug and control} paradigm so that any off-the-shelf political bias classifier\footnote{We assume the classifier employs a common tokenizer.} could be seamlessly integrated.

\paragraph{Logits Projection}
To obtain the almost one-hot logits similar to the initial data distribution, we further project logits $\hat{\mathbf{S}}_{\text{ctr},t}$ at the end of every iteration following \citep{han2023ssd}: 
\vspace{-5pt}
\begin{align}
\mathbf{\hat{S}}_{\text{proj}, t}^{(j)} \text{=} \begin{cases}
+K \text{ if $j$=} \text{top-}p\text{-sampling}(\hat{\mathbf{S}}_{\text{ctr},t})\\
-K \text{ otherwise}
\end{cases} 
\nonumber
\end{align}
where $\text{top-}p$ is the hyperparameter for nucleus sampling \citep{holtzman2019curious}.  After projecting $\mathbf{\hat{S}}_{\text{ctr}, t}$ to $\mathbf{\hat{S}}_{\text{proj}, t}$ , we add a noise according to the forward diffusion schedule and pass the representation $\mathbf{S}_{t}$ as the incoming logits  for the next iteration $t-1$: 
\vspace{-10pt}
\begin{align}
     \mathbf{S}_{t} = \sqrt{\Bar{\alpha}_{t}} \mathbf{\hat{S}}_{\text{proj}, t} + \sqrt{1-\Bar{\alpha}_{t}} \bm{\epsilon}_{t}\nonumber
\end{align}

\vspace{-5pt}

So the decoding process can be summarized as iteratively denoising logits $\mathbf{S}_{T}$ to obtain $\mathbf{S}_{t+1},\mathbf{S}_{t},\dots,\mathbf{S}_{0}$, and $\mathbf{S}_{0}$ is the final summary. At time step $t$, we first mix the noisy logits $\mathbf{S}_{t+1}$ and the model estimation $\hat{\mathbf{S}}_{sc,t+1}$ from  time step $t+1$ (self-conditioning) and obtain a model estimation for step $t$: $\hat{\mathbf{S}}_{sc,t}$. 
Then, we apply the classifier to predict the perspective for the current estimation $\hat{\mathbf{S}}_{sc,t}$ and compare it with a target stance $y$. The difference between the prediction and the target stance is backpropagated to steer the logits $\hat{\mathbf{S}}_{\text{ctr},t}$.  
After that, we project the logits $\hat{\mathbf{S}}_{\text{ctr},t}$ to $\hat{\mathbf{S}}_{\text{proj},t}$ and add Gaussian noise to derive $\mathbf{S}_{t}$. 
Such process is repeated $T$ times with $\mathbf{S}_{0}$ as the final representation. 
The final summary is obtained by converting $\operatorname{argmax}\mathbf{S}_{0}$ to natural language tokens. 
\vspace{-20pt}
\begin{align}
    \mathbf{\hat{S}}_{\text{sc}, t} &= \hat{\mathbf{S}}_{\bm{\theta}}( \frac{  \mathbf{S}_{t+1} + \mathbf{\hat{S}}_{\text{sc}, t+1}  }{2} , t+1)\nonumber \\
    \mathbf{\hat{S}}_{\text{ctr}, t} &= \mathbf{\hat{S}}_{\text{sc}, t} + \lambda \nabla_{\mathbf{\hat{S}}_{\text{sc}, t}} f_{\phi}(y \mid \operatorname{sm}(\mathbf{\hat{S}}_{\text{sc}, t})) 
    \nonumber \\
    \mathbf{\hat{S}}_{\text{proj}, t} &= \operatorname{logits-projection}(\mathbf{\hat{S}}_{\text{ctr}, t})
    \nonumber \\
    \mathbf{S}_{t} &= \sqrt{\Bar{\alpha}_{t}} \mathbf{\hat{S}}_{\text{proj}, t} + \sqrt{1-\Bar{\alpha}_{t}} \bm{\epsilon}_{t}\nonumber
\end{align}

\section{Experiments}
\label{sec:experiment}

\subsection{Experimental Settings}
\begin{table*}[ht]
\centering
 \resizebox{0.8\linewidth}{!}{
 \begin{tabular}{@{}lcccccccl@{}}
 \toprule[1.5pt]
 \multirow{3}{*}{\textbf{Method}} &
 \multirow{3}{*}{\textbf{Pres.}} &
 \multirow{3}{*}{\textbf{Model Size}} &
 \multicolumn{2}{c}{\textbf{POLITICS}} &
 \multicolumn{2}{c}{\textbf{CNN/DM}} &
 \multicolumn{2}{c}{\textbf{XSUM}} \\
 \cmidrule(lr){4-5} \cmidrule(lr){6-7} \cmidrule(lr){8-9}
 &
 &
 &
 \textsc{Suc}$\uparrow$&
 \textsc{Dist}$\downarrow$&
 \textsc{Suc}$\uparrow$&
 \textsc{Dist}$\downarrow$&
 \textsc{Suc}$\uparrow$&
 \textsc{Dist}$\downarrow$\\ \midrule[0.75pt]
 \textsc{t5} &
 \xmark &
 200M &
 44.10 &
 0.35 &
 47.13 &
 0.38 &
 50.53 &
 0.35 \\
 \textsc{brio} &
 \xmark &
 400M &
 44.95 &
 0.35 &
 48.65 &
 0.37 &
 29.19 &
 0.49\\
  \textsc{pegasus} &
 \xmark &
 568M &
 44.19 &
 0.36 &
 44.03 & 
 0.37 &
 25.40 &
 0.51\\
 \textsc{vicuna} &
 \xmark &
 7B &
 52.01 &
 0.30 &
 42.71 &
 0.38 &
 53.19 &
 0.31 \\
 \textsc{falcon} &
 \xmark &
 40B &
 41.51 &
 0.41 &
 40.78 &
 0.39 &
 31.58 &
 0.45\\
  \textsc{llama2} &
 \xmark &
 70B &
 41.97 &
 0.42 &
 43.40 &
 0.39 &
 43.03 &
 0.35 \\
 \midrule
 \textsc{t5} &
 \cmark &
 200M &
 47.29 &
 0.34 &
 41.83&
 0.40 &
 47.97 &
 0.38\\
 \textsc{brio} &
 \cmark &
 400M &
 42.15 &
 0.38 &
 46.98 &
 0.38 &
 30.96 &
 0.48 \\
 \textsc{pegasus} &
 \cmark &
 568M &
 42.38 &
 0.36 &
 43.78 &
 0.38 &
 31.28 &
 0.48 \\
 \textsc{vicuna} &
 \cmark &
 7B &
 53.52 &
 0.29 &
 48.07 &
 0.36 &
 46.02 &
 0.34\\
 \textsc{falcon} &
 \cmark &
 40B &
 39.64 &
 0.42 &
 46.64 &
 0.36 &
 37.63 &
 0.41 \\
 \textsc{llama2} &
 \cmark &
 70B &
 40.15 &
 0.45 &
 43.38 &
 0.44 &
 51.54 &
 \textbf{0.30}\\

 \midrule
 \ourmethod{} 
(ours) &
 \cmark &
 125M &
 \textbf{54.36} &
 \textbf{0.28} &
 \textbf{55.32} &
 \textbf{0.31} &
 \textbf{54.75} &
 0.33\\
   \bottomrule[1.5pt]
\end{tabular}
}
\caption{Performance of political perspective preservation on the three datasets. ``Pres.'' indicates whether the model is instructed to preserve stances or not. $\uparrow$ and $\downarrow$ indicate whether the metric should be high or low.
\ourmethod{} outperforms all baseline models that are 1.6x to 560x larger on five of the six settings across the three datasets.
}
\label{tab:big}
\vspace{-15pt}
\end{table*}
\textbf{Datasets} We adopt three news datasets: CNN/DM \citep{nallapati2016abstractive}, XSUM \citep{narayan2018don}, and POLITICS \citep{liu-etal-2022-politics}. Since there are ground truth labels provided in the POLITICS dataset, we directly employ them to measure the performance of preserving perspectives.

\begin{table}[t]
    \centering
    \resizebox{1\linewidth}{!}{
    \begin{tabular}{@{}lcccccccc@{}}
        \toprule[1.5pt]
         \multirow{2}{*}{\textbf{Method}} &
         \multicolumn{4}{c}{\bf POLITICS} &
         \multicolumn{4}{c}{\bf CNN/DM} \\
         \cmidrule(lr){2-5} \cmidrule(lr){6-9} 
          &
         R1 &
         R2 &
         R-L &
         R-avg &
         R1 &
         R2 &
         R-L &
         R-avg \\ \midrule
         \textsc{t5} &
         38.31 &
         18.04 &
         27.82 &
         33.07 &
         40.82 &
         18.30 &
         28.64 &
         29.25 \\
         \textsc{brio} &
         47.91 &
         24.24 &
         33.12 &
         35.09 &
         46.21 &
         22.04 &
         31.36 &
         33.20 \\
         \textsc{pegasus} &
         40.62 &
         19.36 &
         29.64 &
         29.87 &
         42.70 &
         19.69 &
         29.76 &
         30.72 \\
         \textsc{vicuna} &
         21.33 &
         8.84  &
         14.78 &
         14.98 &
         13.20 &
         3.48  &
         8.51  &
         8.40  \\
         \textsc{falcon} &
         18.77 &
         4.32  &
         11.28 &
         11.46 &
         15.59 &
         3.17  &
         9.43  &
         9.40  \\
         \textsc{llama2} &
         30.93 &
         12.98 &
         20.72 &
         21.54 &
         22.21 &
         6.75  &
         13.89 &
         14.28 \\ \midrule
         \ourmethod{} (ours) &
         37.48 &
         16.50 &
         26.01 &
         26.66 &
         41.12 &
         18.20 &
         27.73 &
         29.02 \\ \bottomrule[1.5pt]     
    \end{tabular}
    }
    \caption{Rouge scores on POLITICS and CNN/DM. Though the decoding process is steered by classifier gradients to preserve political stances, \ourmethod{}'s summarization utility is still competitive among baselines.}
    \label{tab:rouge}
    \vspace{-15pt}
\end{table}

\noindent\textbf{Baselines}
We compare \ourmethod{} with two types of baselines: 1) \emph{summarization systems}, specifically \textsc{BRIO} \citep{liu-etal-2022-brio}, \textsc{Pegasus} \citep{zhang2020pegasus}, and \textsc{T5} \citep{raffel2020exploring}.
2) \emph{large language models}, specifically Vicuna \citep{vicuna2023}, Falcon \citep{penedo2023refinedweb}, and Llama-2 \citep{touvron2023llama}.\footnote{We test them in the zero-shot setting.} For each baseline, we employ two modes: \emph{without preservation}, where the baseline is directly used for summarization; \emph{with preservation}, where we prepend instructions to encourage stance preservation.\footnote{For similar baselines of controllable text generation such as \citet{liu2021dexperts}, we do not compare them with our method since the classifier we use is a discriminator, not a generator as required by the paper.}

\noindent\textbf{Implementation}
We employ the encoder-only \textsc{roberta-base} \citep{liu2019roberta} as the backbone of \ourmethod{}'s diffusion component. To preserve perspectives at inference time, we leverage the political bias classifier from \textsc{POLITICS} \citep{liu-etal-2022-politics}, which measures the political stance of the generation and compares it with the original stance at each decoding step. This allows a loss term measuring the political stance difference to back-propagate to the embedding layers, penalizing perspective inconsistencies. We provide full details of \ourmethod{} training and inference in Appendix \ref{appendix:exp_details}.

\noindent\textbf{Evaluation}
We define two metrics to evaluate the success of preserving political stances in the summary using the political stance classifier that maps text sequences to a bias label $f_{\textit{bias}}(\cdot): \mathrm{str} \rightarrow \{-1, 0, 1\}$ representing left, center, and right-leaning. 1) \emph{\underline{Success Rate}} ($\mathrm{Suc}$): $\frac{1}{|\mathcal{D}|} \sum_{\boldsymbol{d} \in \mathcal{D}} \mathds{1}(f_{\textit{bias}}(\boldsymbol{d}) = f_{\textit{bias}}(\boldsymbol{s}))$, where $\mathds{1}(\cdot)$ denotes the indicator function and $\mathcal{D}$ denotes the full dataset. 2) \emph{\underline{Stance Distance}} ($\mathrm{Dist}$): $\frac{1}{|\mathcal{D}|} \sum_{\boldsymbol{d} \in \mathcal{D}} |f_{\textit{bias}}(\boldsymbol{d}) - f_{\textit{bias}}(\boldsymbol{s})|$. While $\mathrm{Suc}$ examines whether the stance of the summary is consistent with the article, $\mathrm{Dist}$ further evaluates how far the perspective of summaries drifts from the news documents. For summarization utility evaluation, we employ Rouge-1/2/L scores \citep{lin-2004-rouge} and abstractiveness scores \citep{chan-etal-2021-controllable}.






\begin{table*}[t]
    \setlength{\tabcolsep}{15pt}
    \centering
    \resizebox{0.9\linewidth}{!}{
    \begin{tabularx}{\linewidth}{@{} p{6cm}| p{0.5cm}<{\centering} m{4cm}|m{0.3cm}<{\centering}}
        \toprule[1.5pt]
         Context&
         Model&
         Summary&
         Stance \\ \midrule[0.75pt]
         \vspace{-17pt}
         \multirow{3}{6cm}{\small  Biden \dots will confront a divided country beset by an unprecedented and complex set of difficulties \dots Election returns and exit polls revealed sharp differences between men and women and white and minority Americans.\dots His response to these challenges will be limited by a Republican Senate, a solidly conservative Supreme Court majority, hostility from Trump supporters \dots Biden enjoyed a big edge with non-white Americans while white voters stuck with the incumbent\dots (\textbf{center})
        }&
        Ours&
        \small Election returns and exit polls reveal sharp differences between men and women and white.\dots &
        center \cmark\\
         &
         \textsc{t5}&
         \small Biden \dots \textcolor{left}{will be limited by a Republican Senate, a solidly conservative Supreme Court majority, hostility from Trump supporters.} \dots &
         \textcolor{left}{left} \xmark\\
         &
         \textsc{brio}&   
         \small  \dots \textcolor{right}{Biden must confront the pandemic, rebuild the economy and address climate change} \dots &
         \textcolor{right}{right} \xmark\\
         \bottomrule[1.5pt]
    \end{tabularx}
    }
    \caption{A qualitative example of generated summaries from different approaches. Existing summarization systems often alter the political perspective by presenting partial facts or making up non-existing statements. Our method successfully preserves the original perspective by presenting only the main idea and facts in the original article.}
    \label{task:qualitative}
\end{table*}

\subsection{Results}

\textbf{Preserving Author Perspectives}
Table \ref{tab:big} demonstrates that \ourmethod{} achieves the highest average success rate as well as the lowest stance distance across five of the six settings, outperforming baselines that are 1.6x to 560x larger. For success rate, we surpass the second-best method by $1.6\%$, $13.7\%$, and $ 2.9\%$ respectively on the \textsc{politics},\textsc{cnn/dm}, and \textsc{xsum} datasets. This suggests that the combination of diffusion language models and plug-in political bias classifiers offers a promising approach to preserving political perspectives in news summarization.

For large language model baselines that perform text summarization in a zero-shot setting, we observe that adding instructions for stance preservation produces mixed effects on their performance. For example, the instructions work for \textsc{falcon} on CNN/DM but are counterproductive on POLITICS. We hypothesize that large language models struggle to grasp the concept of preserving political opinions off-the-shelf, potentially influenced by their internal notion of political leanings that is often biased and inaccurate \citep{shaikh2022second,feng-etal-2023-pretraining}. However, with an explicit classifier-based gradient steering paradigm, \ourmethod{} successfully advances the ability to preserve political perspectives in generated summaries.

\setlength{\aboverulesep}{0pt}
\setlength{\belowrulesep}{0pt}
\begin{table}[t]
    \centering
    \resizebox{0.8\linewidth}{!}{
   \begin{tabular}{@{}l|ccc@{}}
         Method &
         POLITICS &
         CNN/DM &
         XSUM\\
         \midrule
         \textsc{t5} &
         9.02 &
         8.61 &
         7.15\\
         \textsc{brio} &
         5.17 &
         4.11 &
         3.16\\
         \textsc{pegasus} &
         6.76 &
         3.80 &
         6.46\\
         \textsc{vicuna} &
         3.98 &
         2.64 &
         1.50\\
         \textsc{falcon} &
         1.77 &
         0.83 &
         0.65\\
         \textsc{llama2} &
         3.99 &
         2.20 &
         1.29\\
         \midrule
         \ourmethod{} (ours) &
        6.32 &
        2.59 & 
        2.93\\
    \end{tabular}
    }
    \caption{Abstractiveness scores \citep{chan-etal-2021-controllable}, the lower the better. \ourmethod{} successfully produces concise summaries that are competitive with existing approaches while improving perspective preservation.}
    \label{tab:abstractiveness}
\end{table}

\begin{table}[]
    \centering
    \resizebox{0.4\linewidth}{!}{
    \begin{tabular}{@{}l|c@{}}
         Method & \textsc{cnn/dm}  \\
         \midrule
         \textsc{GPT-Davinci} & 0.8444\\
         \textsc{ChatGPT} & 0.8935\\
         \textsc{Pegasus} & 0.9395\\
         \midrule
         \ourmethod{} (ours) & 0.9289
    \end{tabular}
    }
    \caption{Factuality scores (0-1) of the models measured by FactKB \citep{feng2023factkb}.}
    \label{tab:factuality}
    \vspace{-10pt}
\end{table}

\noindent\textbf{Summarization Utility} We evaluate \ourmethod{} and baselines on CNN/DM and POLITICS by comparing them to reference summaries\footnote{Since no human-written reference summaries are provided in POLITICS, we use LLM-generated summaries selected by humans as reference summaries. To complement the model-generated summaries, we provide an additional human evaluation on the POLITICS dataset in Appendix \ref{appendix:human_eval}.} and present results in Tables \ref{tab:rouge}, \ref{tab:factuality} and \ref{tab:abstractiveness}. Table \ref{tab:rouge} demonstrates that \ourmethod{} achieves Rouge scores that are on-par with state-of-the-art approaches, while Table \ref{tab:abstractiveness} shows that \ourmethod{} is producing abstractive and concise summaries. Table \ref{tab:factuality} suggests that existing approaches and ours are both very factual, echoing the trend of discoveries that summarization systems based on LLMs or their outputs are now overwhelmingly factual, as evaluated by factuality evaluation models\footnote{While their factuality scores are high, we find that existing approaches selectively summarize certain sides of the argument, potentially informed by their internal biases. While the information in the summary indeed appeared and is consistent with the news article, it only presents a partial picture of news events and does not account for the full picture. As a result, the focus of our study is stance preservation, aiming to mitigate this partial summarization issue.}. Together these results demonstrate that \ourmethod{} gets better at preserving political opinions without greatly sacrificing summarization quality.



\noindent\textbf{Qualitative Analysis} In Table \ref{task:qualitative}, we present an example news article from the POLITICS dataset, where models produce summaries with different political leanings.  The original article takes a mostly neutral stance, analyzing the electorate and voter issues. However, \textsc{t5} generates a strongly left-leaning summary by priming the hostility from Republicans and focusing on incorrect facts such as a Republican Senate to support its argument.\footnote{In 2020, Democrats narrowly won control of the senate with a tie-breaking vote from the Vice President.} \textsc{brio} instead makes a right-leaning pitch by highlighting the challenges looming for the incoming administration. In contrast, \ourmethod{} maintains a neutral standpoint, summarizing the demographic differences in the 2020 election and preserving the original article's political stance, as confirmed by the stance classifier.

\section{Analysis and Discussion}
\begin{figure}[t]
	\centering

 \includegraphics[width=0.9\linewidth]{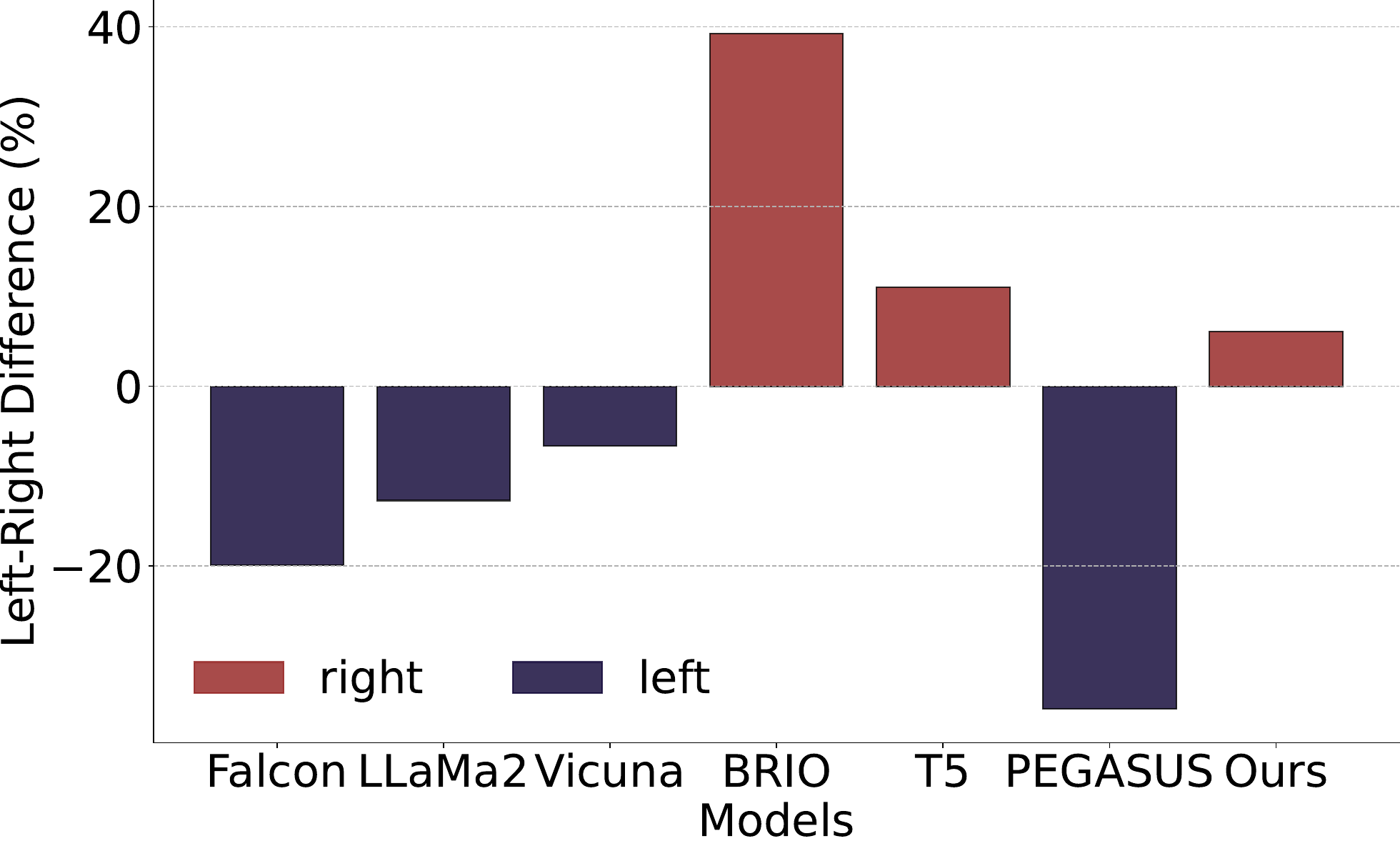}
	\caption{We measure models' inherent biases by averaging the shift in political stances across all center-leaning articles in POLITICS. \ourmethod{} with explicit controllable generation has the lowest absolute bias.}
	\label{fig:inherent}
\end{figure}
\paragraph{Inherent Bias of Models}
Previous works suggest that LLMs could have inherent social and political biases \cite{feng-etal-2023-pretraining,abdulhai2023moral,kurita2019measuring,manzini-etal-2019-black,cheng2023marked,ladhak2023pre}. We now explore how LLM inherent biases could prevent models from preserving author perspectives in news summarization. Given center-leaning articles, we take the summaries generated from different systems and measure their political leaning. We then calculate the difference between the frequency of right-leaning summaries and left-leaning ones for each model and present the results in Figure \ref{fig:inherent}. Baselines such as \textsc{brio} are consistently steering summaries toward the right while most LLMs result in leftward shifts.  We argue that these inherent biases present challenges in preserving political perspectives by reinforcing views from one angle, while \ourmethod{} with specific classifier control has the lowest average bias and mitigates these issues.

\begin{figure}[t]
	\centering

 \includegraphics[width=0.8\linewidth]{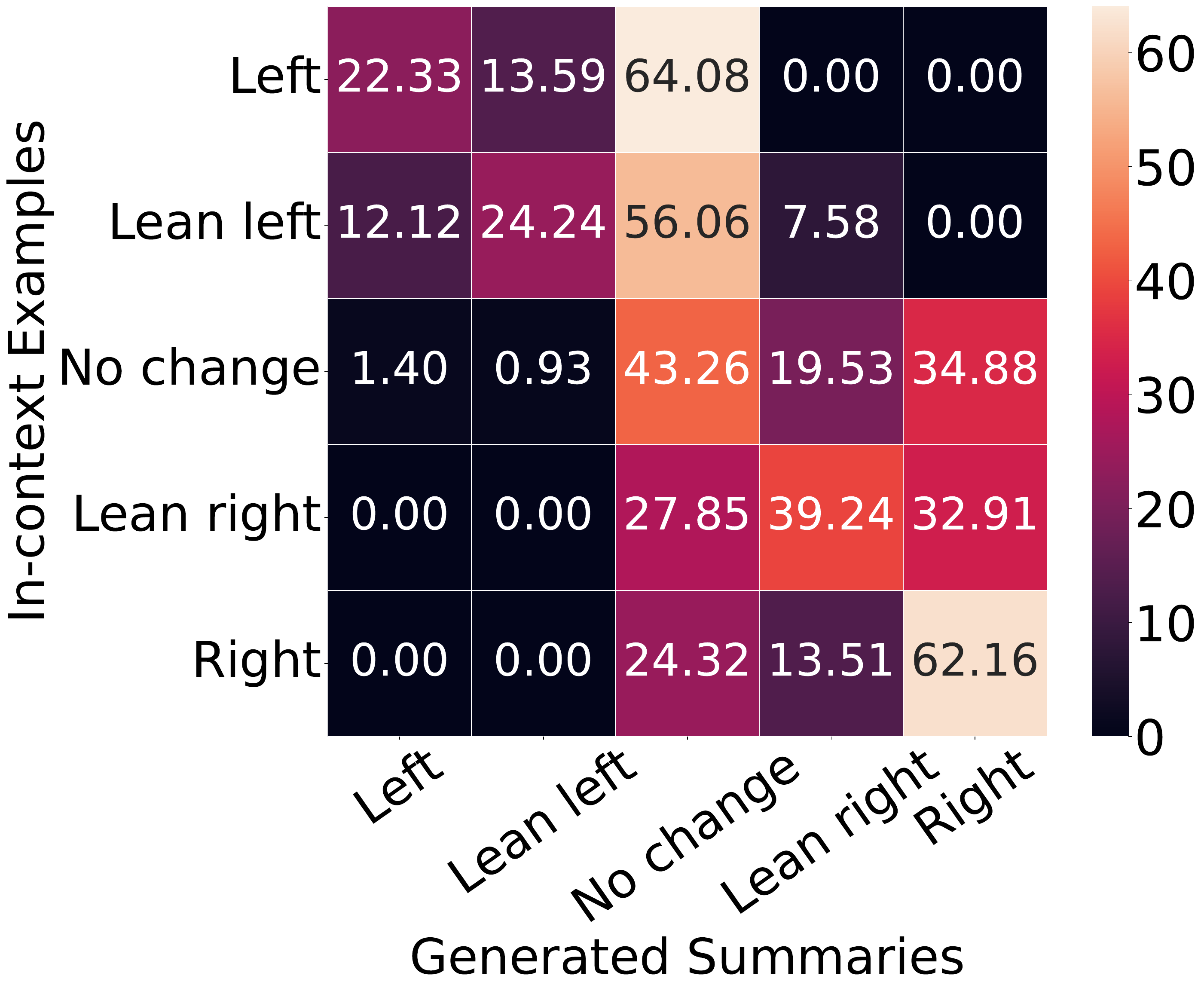}
	\caption{We show how gold summaries as in-context examples alter the perspectives and how model-generated summaries are affected accordingly. We provide \textsc{chatgpt} with both articles and gold summaries as in-context examples. The left-rightward shift of examples can greatly increase the possibility of similar shifts in the model-generated summaries. }
	\label{fig:zero_few_shot}
\end{figure}

\paragraph{Effects of Misleading Gold Summary}
To explore how inconsistent gold summaries can mislead the models, we compare experiments with \textsc{chatgpt} in the few-shot setting shown in Figure \ref{fig:zero_few_shot}. The passage and the corresponding gold summary will be provided first as an example, and then the article will be given again to ask for the model's summary.  We measure how gold summary changes the perspectives of the author and the effects on the model-generated summaries.
It is noteworthy that if a reference summary changes the political leaning toward "right" or "lean right", the chance of \textsc{chatgpt} generating a  "right" or "lean right" summary will be improved. And there is a  similar trend for the left-leaning examples.



\begin{table}[t]
    \centering
    \resizebox{1\linewidth}{!}{
    \begin{tabular}{@{}l|cccccccc@{}}
    \toprule[1.5pt]
         \multirow{2}{*}{\textbf{Ablation}} &
         \multicolumn{2}{c}{\textbf{POLITICS}} &
         \multicolumn{2}{c}{\textbf{CNN/DM}} &
         \multicolumn{2}{c}{\textbf{XSUM}} \\
         \cmidrule(lr){2-3} \cmidrule(lr){4-5} \cmidrule(lr){6-7}
         &
         \textsc{Suc}$\uparrow$&
         \textsc{Dist}$\downarrow$&
         \textsc{Suc}$\uparrow$&
         \textsc{Dist}$\downarrow$&
         \textsc{Suc}$\uparrow$&
         \textsc{Dist}$\downarrow$\\ \midrule[0.75pt]
         \ourmethod{} &
          54.36 &
          0.56 &
          55.32 &
          0.62 &
          54.75 &
          0.65 \\ \midrule[0.75pt]
          w/o MC &
          33.66 &
          0.93 &
          39.53 &
          0.81 &
          52.44&
          0.69 \\
          \textit{change} &
          \textit{-20.70} &
          \textit{+0.37} &
          \textit{-15.79} &
          \textit{+0.19} &
          \textit{-2.31} &
          \textit{+0.04}\\
          w/o SC &
          47.36 &
          0.65 &
          44.61 &
          0.78 &
          45.95 &
          0.70\\
          \textit{change} &
          \textit{-7.00} &
          \textit{+0.09} &
          \textit{-10.71} &
          \textit{+0.16} &
          \textit{-8.80} &
          \textit{+0.05}\\
          \bottomrule[1.5pt]
    \end{tabular}
    }
    \caption{Ablation study investigating how modular control (MC) and self-conditioning (SC) contribute to \ourmethod{}'s performance.}
    \label{tab:ablation}
\end{table}

\paragraph{Ablation Study}
We observe how \ourmethod{}'s performance degrades by dropping the modular control (MC) or self-conditioning (SC) and present the results in Table \ref{tab:ablation}. It is shown that modular control has a significant impact on forcing the model to be faithful to the original opinions. The preserving capacity also drops without self-conditioning.

\section{Related Work}
\paragraph{Text Summarization and Factuality Evaluation}
Research on neural text summarization has produced models and systems that are capable of generating fluent and informative summaries \citep{liu2019text, balachandran2021structsum, rothe2021thorough, narayan2021planning, bhattacharjee2023crosssum, chen2023unisumm, he-etal-2023-z, liu-etal-2023-revisiting, chen-etal-2023-improving-robustness}, given documents from various domains such as news articles \citep{fabbri-etal-2019-multi, liu-etal-2022-end, bahrainian-etal-2022-newts}, scientific literature \citep{goldsack2022making}, social media and dialogue \citep{tang2022confit, liu2022data}. However, it remains challenging to generate summaries that are factually consistent with the given document \citep{cao2018faithful, balachandran2022correcting}, resulting in the research area of factuality evaluation. Existing works propose benchmarks to evaluate the factuality of generated summaries \citep{pagnoni-etal-2021-understanding, tang-etal-2023-understanding}, develop factuality evaluation models and metrics \citep{wang2020asking,kryscinski-etal-2020-evaluating,nan2021improving,goyal2021annotating, ribeiro2022factgraph,utama2022falsesum,laban2022summac, feng2023factkb,luo2023chatgpt}, and improve the factuality of generated summaries \citep{aharoni2023multilingual, liu-etal-2023-improving}. Recent studies suggest that state-of-the-art large language models \citep{goyal2022news, bhaskar2022zero} are capable of achieving remarkable factuality in text summarization. However, while LLMs are capable of generating summaries that are factually faithful, our work demonstrates that they struggle to generate summaries that are faithful to the authors' original opinions and perspectives (Figure \ref{teaser}). As a result, we propose \ourmethod{}, an important first step towards summarization systems that preserve the authors' opinions in the generated summary.
\vspace{-5pt}
\paragraph{Understanding the Social and Political Biases of Language Models}
Extensive research has demonstrated that machine learning models could encode and exhibit social and political biases \citep{ bender2021dangers, jin2021transferability, shaikh2022second, li-etal-2022-herb}. Existing works mainly analyze biases expressed in word embeddings \citep{bolukbasi2016man, caliskan2017semantics, kurita2019measuring}, token probabilities \citep{borkan2019nuanced, bordia-bowman-2019-identifying, liu2021mitigating}, model performance discrepancy \citep{hardt2016equality, feng-etal-2023-pretraining}, and generated texts \citep{kumar2022language}. Specifically for political biases, several studies have been proposed to probe LLMs \citep{bang2021assessing,feng-etal-2023-pretraining}, evaluate the political leaning of texts \citep{feng2021kgap, zhang-etal-2022-kcd, liu-etal-2022-politics, qiu-etal-2022-late}, and pretraining LMs on partisan corpora \citep{jiang-etal-2022-communitylm}. Annotator \citep{sap2019risk, sap2022annotators, gordon2022jury} and data bias \citep{dixon2018measuring, dodge2021documenting, harris2022exploring} are commonly attributed as the cause of LM biases, while existing works also established that LM biases could propagate into downstream tasks and cause fairness issues \citep{ li2020unqovering, feng-etal-2023-pretraining, steed2022upstream,ladhak2023pre}. In this work, we uniquely focus on the task of news summarization: while existing LM-based summarization approaches generate summaries being inconsistent with the political stances of the article, we propose \ourmethod{} to steer the perspective of the summary through iterative controllable generation.


\vspace{-5pt}
\paragraph{Controllable Text Generation}
In text summarization, controllable text generation can generate summaries with given entities, predefined lengths, and more \citep{chan-etal-2021-controllable,he2020ctrlsum,li2022diffusion}. 
More generally, inference-time methods can be used to steer the generation process by altering the output probability distribution at decoding time \citep{dathathri2019plug,krause2021gedi,yang2021fudge,liu2021dexperts,lu-etal-2021-neurologic,pascual-etal-2021-plug-play,kumar2021controlled,Qin2022COLDDE,kumar-etal-2022-gradient,mireshghallah2022mix}. 
Particularly, \citet{han-etal-2023-ssd} leverage diffusion-based methods that apply inference-time control through off-the-shelf classifiers. 
In this work, we further explore the summarization setup using diffusion models to preserve opinions in the decoding process. 
\vspace{-5pt}
\section{Conclusion}
\vspace{-5pt}
We demonstrate that existing summarization systems and LLMs struggle to preserve the authors' political perspectives in news summarization. We present \ourmethod{}, a diffusion-based summarization model that improves political perspective preservation by iteratively guiding the decoding process with an external political stance classifier. Extensive experiments demonstrate that \ourmethod{} outperforms large language models and summarization systems in producing summaries faithful to the political stances of news documents while maintaining competitive summarization utility.

\section*{Limitations}
\paragraph{Trade off between Utility and Preservation} While \ourmethod{} has achieved state-of-the-art performance in preserving author perspectives among all methods, steering the stance during the inference time can affect the utility of the summary, which results in lower rouge scores or abstractiveness measures. As shown in Figure \ref{tab:gold}, the gold summaries provided in the datasets do have biases and not the ideal references for preserving original perspectives, which motivates this work and future directions to improve model stability in controllable summarization.
\paragraph{Time Overhead} Diffusion models for language are notoriously slower at inference time. While our proposed \ourmethod{} is better than existing summarization systems and LLMs at preserving authors' political perspectives in the generated summaries, it comes at the cost of inference time subject to the classifier control component at the decoding time of diffusion models. We employ 1000 decoding steps to refine a generated summary so that it is consistent with the news articles' perspectives and stances, which adds to inference-time computational costs.

\paragraph{Political Bias Classifier} We employ POLITICS \citep{liu-etal-2022-politics}, an LM-based political bias classifier to iteratively steer the political stances of the generated summary. While it successfully helps to preserve author perspectives, it only provides coarse-grained categorical political leanings (left/center/right). Besides, it is shown in \citet{liu-etal-2022-politics} that this political bias classifier is not $100\%$ accurate at identifying political stances, which may mislead the process of preserving the original opinions. Besides, since the classifier we use is based on American political news sources, the political leanings defined in this paper are according to the US policy. There will be different definitions for other countries.  However, we argue that our proposed methodology in  \ourmethod{} is general and compatible with future political bias classifiers that are more fine-grained, accurate, and appropriate.

\section*{Ethics Statement}
Although \ourmethod{}'s intended use case is to preserve author perspectives in news summarization, there is a potential risk for misuse of controllable generation models: the same methodology can be used to steer the political leaning of the generated summary towards the hyperpartisan extremes, furthering societal divides and deepening polarization. Therefore, we plan to establish access permission to the fine-tuned \ourmethod{} weights to ensure that it is only used for research purposes. 

\section*{Acknowledgements}
This research is supported in part by the Office of the Director of National Intelligence (ODNI), Intelligence Advanced Research Projects Activity (IARPA), via the HIATUS Program contract \#2022-22072200004. 
This material is also funded by the DARPA Grant under Contract No.~HR001120C0124. 
We also gratefully acknowledge support from NSF CAREER Grant No.~IIS2142739, NSF Grants No.~IIS2125201, IIS2203097, and the Alfred P.~Sloan Foundation Fellowship.
The views and conclusions contained herein are those of the authors and should not be interpreted as necessarily representing the official policies, either expressed or implied, of ODNI, IARPA, or the U.S. Government. The U.S.~Government is authorized to reproduce and distribute reprints for governmental purposes notwithstanding any copyright annotation therein.

\bibliography{ours}

%

\appendix

\section{Changes in Political Stances between Model-Generated Summaries and the Articles}
\label{changes_cont}
\begin{table*}[t]
    \centering
    \resizebox{0.7\linewidth}{!}{
    \begin{tabular}{@{}lcccccc@{}}
        \toprule[1.5pt]
        \textbf{\textsc{change}}&  
        \textbf{\textsc{text-davinci}}& 
        \textbf{\textsc{chatgpt}}&
        \textbf{\textsc{pegasus}}&
        \textbf{\textsc{bart}}&
        \textbf{\textsc{brio}}&
        \textbf{\textsc{t5}}\\
        \midrule[0.75pt]
         Left&
         14.4&
         7.6&
         17.0&
         3.8&
         4.2&
         5.0\\
         Lean left&
         8.8&
         9.8&
         14.4&
         9.8&
         8.0&
         8.2\\
         Center&
         44.2&
         41.2&
         45.0&
         40.6&
         39.0&
         38.4\\
         Lean right&
         14.0&
         18.2&
         18.6&
         23.4&
         15.6&
         20.6\\
         Right&
         18.6&
         27.0&
         5.0&
         18.6&
         33.2&
         27.8\\       
         \bottomrule[1.5pt]
    \end{tabular}
    }
    \caption{Changes in political stances between the generated summaries and the news article. }
    \label{tab:model_gen}
\end{table*}
We present specific numbers describing how model-generated summaries shift the author's points of view in Table \ref{tab:model_gen}, as a complement for Figure \ref{teaser}.

\begin{table}[t]
    \setlength{\tabcolsep}{15pt}
    \centering
    \resizebox{0.9\linewidth}{!}{
    \begin{tabularx}{\linewidth}{@{}m{0.3cm}<{\centering}| m{6cm}}
        \toprule[1.5pt]
         Pres.&
         Prompt\\ \midrule[0.75pt]
         \vspace{-17pt}
         \xmark &
         \small You are an abstractive summarizer that follows the output pattern. Please write a summary for the document. Document: \\ \midrule[0.75pt]
         \cmark &
         \small You are an abstractive summarizer that follows the output pattern. You should always respect the political bias in the original article.Please write a summary for the document. Document: \\
         \bottomrule[1.5pt]
    \end{tabularx}
    }
    \caption{Prompts for large language models, where "Pres." is the abbreviation for whether it belongs to the preservation setting.}
    \label{task:prompt}
\end{table}
\begin{table}[]
    \centering
    \resizebox{0.8\linewidth}{!}{
    \begin{tabular}{ccc}
    \toprule[1.5pt]
         Method&
         POLITICS&
         CNN/DM\\\midrule[0.75pt]
         \textsc{text-davinci} &
         74.93 &
         84.44 \\
         \textsc{chatgpt} &
         96.15 &
         89.35\\ \bottomrule[1.5pt]    
    \end{tabular}
    }
    \caption{Factuality scores for LLM-generated summaries.}
    \label{tab:fact}
\end{table}

\begin{table}[t]
\centering
\begin{tabular}{@{}ll@{}}
\toprule[1.5pt]
\textbf{Hyperparameter}
& \textbf{Value}\\ \midrule[0.75pt]
training steps & 20000\\
learning rate & $3 \times 10^{-5}$\\
decoding steps & 1000 \\
max target length & 120\\
control learning rate $\lambda$ & 4000\\
simplex value $K$ & 5\\ \bottomrule[1.5pt]
\end{tabular}
\caption{Hyperparamters for \ourmethod{}}
\label{tab:hyperparameter}
\end{table}

\section{Experiment Details}
During fine-tuning on summarization, we use a leaning rate of $3e-5$,  We fine-tuned for 20000 steps. 

For decoding, we use $\text{top-}$  $p=0.95$ suggested in \citet{han-etal-2023-ssd} and $1000$ diffusion steps according to \citet{mahabadi2023tess}.

We implement \ourmethod{} on a server using Tesla V100 GPU with 32 GB memory, 16 CPU cores, and 377GB memory for the experiments. 

The backbone of our model is \textsc{roberta-base}.
It's noticeable that both \ourmethod{} and the model in \citep{liu-etal-2022-politics} use \textsc{roberta-base}, and thus they share the same tokenizer. Therefore, as mentioned in \citep{han-etal-2023-ssd}, they can be used for control in an off-the-shelf manner.

For POLITICS, there are no human-written summaries. Therefore, we take the summarization of \textsc{gpt-turbo} as the ground truth. The details are in the appendix \ref{appendix:select}   

With \textsc{CNN/DM} as a popular dataset in text summarization, we aim to test how well \ourmethod{} can perform traditional summarization tasks. However, not all the news articles in the \textsc{CNN/DM} are within the political discipline, which is inappropriate for political leaning preservation. Therefore, we leverage the \textsc{POLITICS} dataset\citep{liu-etal-2022-politics}, which consists of political news with labels of political leaning.

\label{appendix:exp_details}

\begin{figure}[ht]
	\centering

 \includegraphics[width=1\linewidth]{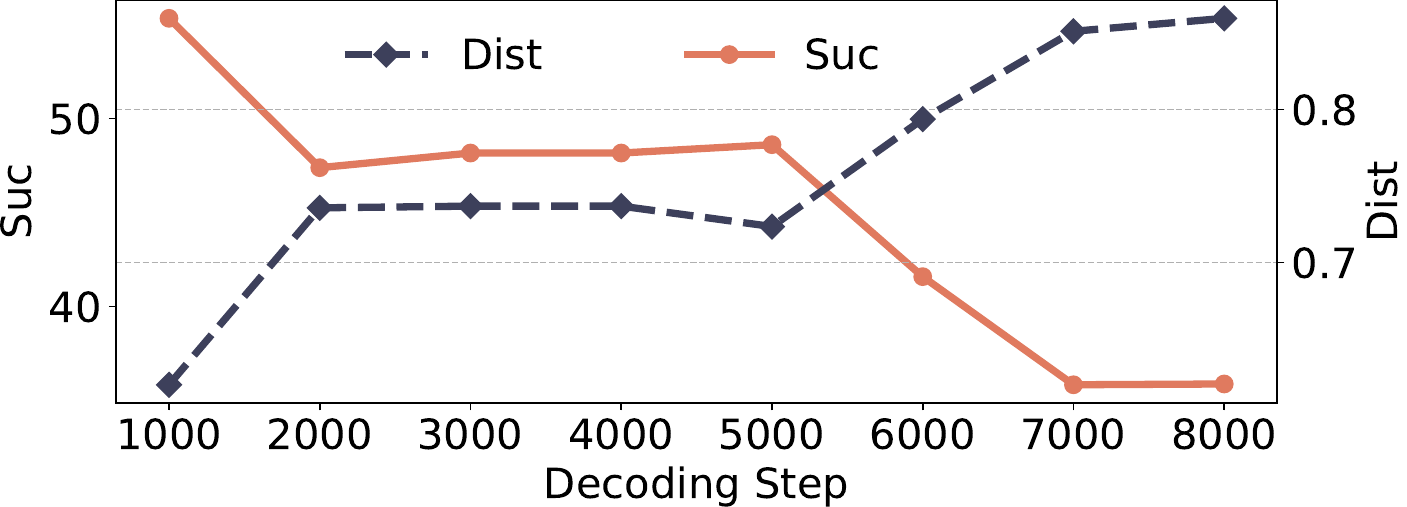}
	\caption{We observe how our model behaves if the total diffusion steps change from $1000$ to $8000$. If the number of total steps is increased beyond $1000$, a drop in the performance would be observed.  }
	\label{fig:d_step}
\end{figure}
\section{Number of Decoding Steps} 

Besides control learning rate, another important hyperparameter is the number of decoding steps in the inference time, which can vary from $1000$ to $5000$ in existing diffusion language models\citet{han-etal-2023-ssd,mahabadi2023tess}. Thus, we observe how our model behaves if the total diffusion steps change from $1000$ to $8000$ and present the results in Figure \ref{fig:d_step}. It is shown that the best performance is achieved at step $=1000$, and gradually drops when the number of decoding steps increases.

\section{Stance Control Learning Rate}
\begin{figure}[t]
	\centering

 \includegraphics[width=1\linewidth]{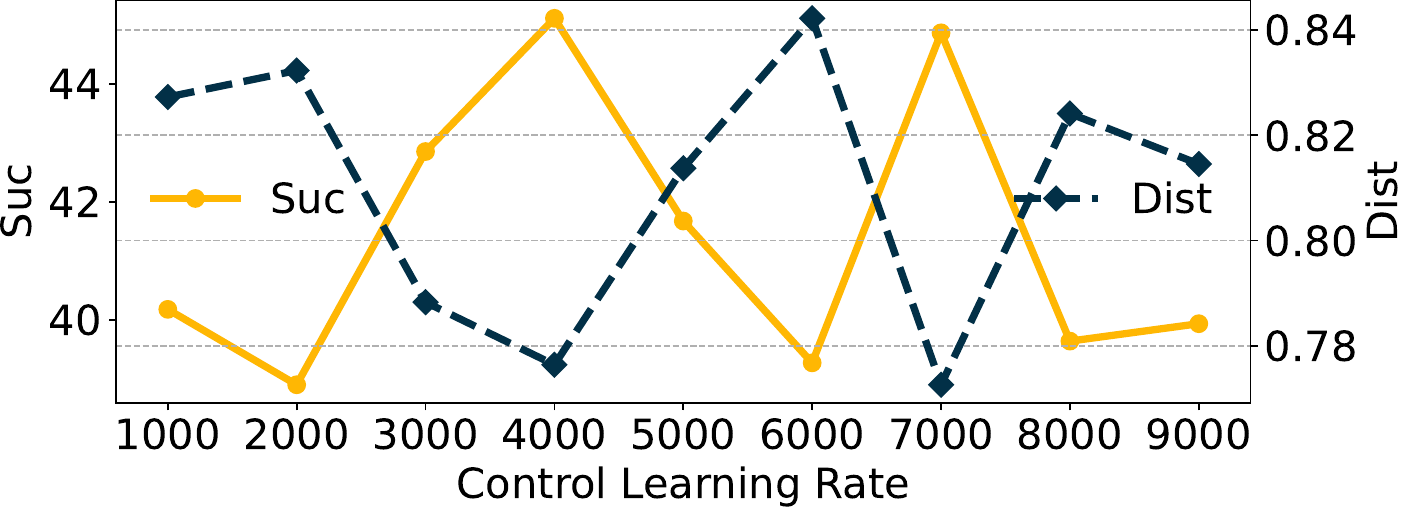}
	\caption{We show how the stance control learning rate $\lambda$ affects model performance. ``Suc'' should be high and ``Dist'' should be low. Best stance preservation is achieved at $\lambda=4000$, while text degeneration happens with higher $\lambda$s.}
	\label{fig:ctr}
         \vspace{-15pt}
\end{figure}

An important hyperparameter in \ourmethod{} is the classifier control learning rate $\lambda$ in equation \ref{equ:ctr}, which determines the intensity of stance steering by controlling the gradients. We show how this parameter affects the model's performance in Figure \ref{fig:ctr}. It is observed that the highest success rate and the lowest distance are achieved at $\lambda=4000$, and the controlling capability then gradually declines when $\lambda$ increases, potentially due to top-$p$ setting \citep{han-etal-2023-ssd}.

\section{Understanding Political Instructions in Prompts}
The prompt we use for zero-shot inference for large language models are listed in the Table \ref{task:prompt}

\section{Human Evaluation}
\label{appendix:human_eval}

To complement the manually selected model summaries and the original labels in the POLITICS dataset, we conduct an additional human evaluation to measure the stance preservation and factuality.150 news articles (50 each for left, center, and right) and their generated summaries are selected from the POLITICS dataset and manually evaluated. The percentage of generated summaries that are faithful to the political stances and facts in the original news article is presented. The results in the Table \ref{tab:human} suggests that human evaluation produces similar results to previous experiments. For example, the average stance preservation rate for our approach is 54.36, while the average with human evaluation is 59.09. This indicates that the current automatic evaluation is a sound solution for scaling such experiments, while we complement it with human evaluation and analysis.

\begin{table*}[h]
    \centering
    \resizebox{0.7\linewidth}{!}{
    \begin{tabular}{@{}lcccccccc@{}}
    \toprule[1.5pt]
         \multirow{2}{*}{Method} &
         \multicolumn{4}{c}{text-davinci as gold} &
         \multicolumn{4}{c}{ChatGPT as gold}\\
         &
         R-1 &
         R-2 &
         R-L &
         R-avg &
         R-1 &
         R-2 &
         R-L &
         R-avg \\ \midrule[0.75pt]
         \textsc{t5} &
         28.40 &
         11.20 &
         21.66 &
         20.42 &
         36.35 &
         17.50 &
         27.62 &
         27.16 \\
         \textsc{brio} &
         31.11 &
         13.66 &
         23.25 &
         22.67 & 
         47.91 &
         24.24 &
         33.12 &
         35.09\\
         \textsc{pegasus} &
         26.10 &
         9.40  &
         19.37 &
         18.29 &
         40.62 &
         19.36 &
         29.64 &
         29.87 \\
         \bottomrule[1.5pt]   
    \end{tabular}
    }
    \caption{Comparison of rouge scores using \textsc{text-davinci} or \textsc{chatgpt} as gold summaries.}
    \label{tab:rouge_comp}
\end{table*}

\begin{table*}[]
    \centering
    \resizebox{0.5\linewidth}{!}{
    \begin{tabular}{@{}l|cccc@{}}
    Category & Left & Center & Right & Avg. \\ \midrule[0.75pt]
    Stance Preservation &63.27 & 56.00 & 58.00 & 59.09 \\
    Factuality &48.98 & 44.00 & 55.00 & 49.33 \\
    \end{tabular}
    }
    \caption{Human evaluation results for \ourmethod{}.}
    \label{tab:human}
\end{table*}

\section{Ablation Study (cont.)}
In addition to  success rate and distance, we also present the results of rouge scores for the ablation settings in Table \ref{tab:ablation_cont}.

\begin{table*}[t]
    \centering
    \resizebox{1\linewidth}{!}{
    \begin{tabular}{@{}l|cccccccccccc@{}}
    \toprule[1.5pt]
         \multirow{2}{*}{\textbf{Ablation}} &
         \multicolumn{4}{c}{\textbf{POLITICS}} &
         \multicolumn{4}{c}{\textbf{CNN/DM}} &
         \multicolumn{4}{c}{\textbf{XSUM}} \\
        \cmidrule(lr){2-5} \cmidrule(lr){6-9} \cmidrule(lr){10-13}
         &
         R-1 &
         R-2 &
         R-L &
         R-avg&
         R-1 &
         R-2 &
         R-L &
         R-avg&
         R-1 &
         R-2 &
         R-L &
         R-avg\\
         \midrule[1.5pt]
         \ourmethod{}&
         37.48 &
         16.50 &
         26.01 &
         26.66 &
         41.12 &
         18.20 &
         27.73 &
         29.02 &
         19.19 &
         2.77 &
         13.08 &
         11.68\\
         w/o MC &
         36.24 &
         16.21 &
         25.58 &
         26.01 &
         39.66 &
         17.52 &
         27.71 &
         28.29 &
         18.51 &
         2.89 &
         12.35 &
         11.25\\
         \textit{change}&
         \textit{-1.24}&
         \textit{-0.29}&
         \textit{-0.43}&
         \textit{-0.65}&
         \textit{-1.46}&
         \textit{-0.69}&
         \textit{-0.02}&
         \textit{-0.72}&
         \textit{-0.68}&
         \textit{0.12}&
         \textit{-0.73}&
         \textit{-0.43}\\
         w/o SC &
         32.60 &
         11.78 &
         21.90 &
         22.09 &
         37.46 &
         13.70 &
         24.89 &
         25.35 &
         19.01 &
         2.53 &
         12.78 &
         11.44 \\
         \textit{change}&
         \textit{-4.88}&
         \textit{4.72}&
         \textit{-4.11}&
         \textit{-4.57}&
         \textit{-3.66}&
         \textit{-4.50}&
         \textit{-2.84}&
         \textit{-3.67}&
         \textit{-0.18}&
         \textit{-0.24}&
         \textit{-0.30}&
         \textit{-0.24}\\
         \bottomrule[1.5pt]

    \end{tabular}
    }
    \caption{Ablation study (cont.) investigating how modular control (MC) and self-conditioning (SC) contribute to \ourmethod{}'s performance.}
    \label{tab:ablation_cont}
    \vspace{-20pt}
\end{table*}

\section{Qualitative Analysis (cont.)}
 Although \ourmethod{} achieves the highest performance on the datasets, it can also fail in certain cases. We present one failure in Table \ref{task:qualitative_cont} and more examples in the following tables.

\section{Selecting Criteria}
\label{appendix:select}
Because there aren't gold summaries in the \textsc{POLITICS}\citep{liu-etal-2022-politics} dataset, we use model-generated summaries for calculating rouge scores.
We prompt the \textsc{text-davinci} and \textsc{chatgpt}, and compare factuality and overall rouge scores.

We calculate the factuality score of summaries by \citet{feng2023factkb} and present the scores in Table \ref{tab:fact}. It is shown that  \textsc{chatgpt} has a higher level of faithfulness.

Choosing \textsc{text-davinci} and \textsc{chatgpt} as reference summaries respectively, we calculate the rouge scores respectively on \textsc{politics} dataset and present the results in Table \ref{tab:rouge_comp}.

We can see that most models achieve higher rouge scores when selecting \textsc{chatgpt} to generate gold summaries, which implies a higher agreement.

\begin{table*}[t]
    \setlength{\tabcolsep}{15pt}
    \centering
    \resizebox{0.9\linewidth}{!}{
    \begin{tabularx}{\linewidth}{@{} p{6cm}| p{0.3cm}<{\centering} p{5cm}|m{0.1cm}<{\centering}}
        \toprule[1.5pt]
         Context&
         Model&
         Summary&
         Stance \\ \midrule[0.75pt]
         \vspace{-15pt}
         \multirow{3}{6cm}{\small  For months, Republican leaders have been uniform in their insistence that they would allow everyone’s taxes to rise if the rich did not get to keep their Bush-era tax breaks. Mr. Obama has proposed continuing the tax cut for the 98 percent of taxpaying families \dots Republicans have demanded tax cuts for all, and, so far, not a single Republican leader has lined up behind Mr. Boehner’s concession. Ultimately, the case for the top-level tax cuts is increasingly shaky. If Republicans are the least bit serious about reducing the deficit, they have to acknowledge that doing so requires additional revenues\dots  (\textbf{left})
        }&
        Ours&
        \small Republican leaders have been ready to maintain Bush-era tax breaks to continue tax rates. Mr. Obama, who has earned less than \$250,000, will keep up with extra revenue at top rates. \dots &
        right \xmark\\
         &
         \textsc{t5}&
         \small The case for the top-level tax cuts is increasingly shaky. \textcolor{left}{If Republicans are the least bit serious about reducing the deficit, they have to acknowledge that doing so requires additional revenues.} &
         \textcolor{left}{left} \cmark\\
         &
         \textsc{brio}&   
         \small \dots Republicans have demanded tax cuts for all, \dots \textcolor{left}{If Republicans are serious about reducing the deficit, they have to acknowledge that doing so requires additional revenues.}&
         \textcolor{left}{left} \cmark\\
         \bottomrule[1.5pt]
    \end{tabularx}
    }
    \caption{Example \#1 of one news article, three summaries generated by \ourmethod{} and two baselines, as well as their stances as evaluated by the political bias classifier.}
    \label{task:qualitative_cont}
    \end{table*}


\begin{table*}[t]
    \setlength{\tabcolsep}{15pt}
    \centering
    \vspace{-20pt}
    \resizebox{0.9\linewidth}{!}{
    \begin{tabularx}{\linewidth}{@{} p{6cm}| p{0.3cm}<{\centering} p{5cm}|m{0.1cm}<{\centering}}
        \toprule[1.5pt]
         Context&
         Model&
         Summary&
         Stance \\ \midrule[0.75pt]
         \vspace{20pt}
         \multirow{3}{6cm}{\small  Biting his nails nervously, these are the first pictures of the migrant boat captain accused of killing 900 men, women and children in one of the worst maritime disasters since World War Two. Tunisian skipper Mohammed Ali Malek, 27, was arrested when he stepped onto Sicilian soil last night, some 24 hours after his  boat capsized in the Mediterranean. Before leaving the Italian coastguard vessel, however, he was forced to watch the bodies of 24 victims of the tragedy being carried off the ship for burial on the island of Malta. He was later charged with multiple manslaughter, causing a shipwreck and aiding illegal immigration. Prosecutors claim he contributed to the disaster by mistakenly ramming the overcrowded fishing boat into a merchant ship that had come to its rescue. \dots  
  (\textbf{left})
        }& 
        Ours&
        \small "Mohammed Ali Malek, 27, was arrested when he stepped into Sicilian soil last night.\dots
        Mamud Bikhit, 26, also charged with 'aiding illegal immigration' Migrants shifted position on the boat, causing it to overturn as result of collision." &
        \textcolor{left}{left} \cmark\\
         &
         \textsc{falcon}&
         \small The EU's foreign chief, Federica Mogherini, called for 'rapid, concrete and coordinated' action to deal with the Mediterranean migrant disaster \dots She said the situation was a 'humanitarian emergency' and urged EU governments to step up their efforts to help refugees \dots &
         \textcolor{right}{right} \xmark\\
         &
         \textsc{llama2}&   
         \small \dots The tragedy has sparked widespread outrage and calls for action from politicians and religious leaders. Pope Francis, who has been an outspoken advocate for migrant rights, said he was'shocked' by the tragedy and called for an end to'such slaughter'.\dots &
         \textcolor{left}{left} \cmark\\
         \bottomrule[1.5pt]
    \end{tabularx}
    }
    \caption{Example \#2 of one news article, three summaries generated by \ourmethod{} and two baselines, as well as their stances as evaluated by the political bias classifier.}
\end{table*}

\begin{table*}[t]
    \vspace{-10pt}
    \setlength{\tabcolsep}{15pt}
    \centering
    \resizebox{0.9\linewidth}{!}{
    \begin{tabularx}{\linewidth}{@{} p{6cm}| p{0.3cm}<{\centering} p{5cm}|m{0.1cm}<{\centering}}
        \toprule[1.5pt]
         Context&
         Model&
         Summary&
         Stance \\ \midrule[0.75pt]
         \multirow{3}{6cm}{\small  At least 34 people were arrested after hundreds of protesters gathered in New York City on Tuesday night to march against police brutality. The march was organized by National Actions to Stop Murder By Police. Many of the protesters cited the deaths of \dots The protesters marched from Manhattan's Union Square and across the Brooklyn Bridge where they partially blocked traffic. Scroll down for video. Hundreds of protesters gathered in New York City on Tuesday night to march against police brutality. \dots Organizers say Tuesday's protest was one of 28 being held across the country. Protesters tried to gain accress to the roadways of the Brooklyn Bridge, but they were blocked by police. \dots Stephen Davis, the Police Department’s chief spokesman, said 34 people had been arrested by 6:40 p.m, reports the New York Times. Police say an off-duty police officer driving home on the bridge was assaulted by two protesters when he got out of his vehicle to investigate. Police say the suspects ran off after he identified himself as a police officer. He was hospitalized with injuries to his face and arm.  \dots (\textbf{left})
        }& 
        Ours&
        \small The march was organized by National Actions to Stop Murder By Police. Protesters cited the deaths of \dots. Protesters marched from Manhattan's Union Square \dots where they partially blocked traffic. Protesters tried to gain accress to roadways but were blocked by police. \dots &
        left \cmark\\
         &
         \textsc{llama2}&
         \small The protesters marched from Manhattan's Union Square and across the Brooklyn Bridge where they partially blocked traffic. \dots Scroll down for video. The protesters marched from Manhattan's Union Square and across the Brooklyn Bridge where they partially blocked traffic. \dots &
         center \xmark\\
         &
         \textsc{brio}&   
         \small The march was organized by National Actions to Stop Murder By Police.\dots  Many of the protesters cited the deaths of Eric Garner in Staten Island and Walter Scott in South Carolina. Police say an off-duty police officer was assaulted by two protesters on the bridge.&
         \textcolor{left}{left} \cmark\\
         \bottomrule[1.5pt]
    \end{tabularx}
    }
    \caption{Example \#3 of one news article, three summaries generated by \ourmethod{} and two baselines, as well as their stances as evaluated by the political bias classifier.}
\end{table*}

\begin{table*}[t]
    \vspace{-10pt}
    \setlength{\tabcolsep}{15pt}
    \centering
    \resizebox{0.9\linewidth}{!}{
    \begin{tabularx}{\linewidth}{@{} p{6cm}| p{0.3cm}<{\centering} p{5cm}|m{0.1cm}<{\centering}}
        \toprule[1.5pt]
         Context&
         Model&
         Summary&
         Stance \\ \midrule[0.75pt]
         \vspace{-15pt}
         \multirow{3}{6cm}{
         \small In Iowa, Ryan says budget a step toward GOP unity. CEDAR RAPIDS, Iowa (AP) — Republican U.S. Rep. Paul Ryan told an Iowa audience Friday that his party can and must come together, and he held out his recently passed budget plan as a sign of growing GOP unity. Although blocs of Republicans object to aspects of the plan passed Thursday in the U.S. House, Ryan said it embodies the principles upon which the nation was founded. “Some people wanted to go further, some people thought it went too far. The point is we unified around these common principles in a plan,” the Wisconsin congressman told reporters after headlining a state party dinner in Cedar Rapids. “That’s very important to me — which is we can’t just oppose, we have to propose.” Ryan, the 2012 Republican vice presidential nominee, also played down the significance of his speech in Iowa \dots (\textbf{center})
        }& 
        Ours&
        \small U.S. Paul Ryan says his party can and must come together.
        Ryan says budget plan a step toward GOP unity.
        Ryan: "Very important to me is we can't just oppose, we have to propose" &
        center \cmark\\
         &
         \textsc{falcon}&
         \vspace{42pt}
         \small \_\_ Follow David Pitt on Twitter at \dots 
         \vspace{42pt}&
          \xmark\\
         &
         \textsc{vicuna}&   
         \small 2014 The Associated Press. All rights reserved. This material may not be published, broadcast, rewritten, or redistributed. &
          \xmark\\
         \bottomrule[1.5pt]
    \end{tabularx}
    }
    \caption{Example \#4 of one news article, three summaries generated by \ourmethod{} and two baselines, as well as their stances as evaluated by the political bias classifier.}
\end{table*}

\begin{table*}[t]
    \vspace{-20pt}
    \setlength{\tabcolsep}{15pt}
    \centering
    \resizebox{0.9\linewidth}{!}{
    \begin{tabularx}{\linewidth}{@{} p{6cm}| p{0.3cm}<{\centering} p{5cm}|m{0.1cm}<{\centering}}
        \toprule[1.5pt]
         Context&
         Model&
         Summary&
         Stance \\ \midrule[0.75pt]
         \vspace{2pt}
         \multirow{3}{6cm}{
         \small Chamber of Commerce hits Democratic Senate candidates on union bill. The Chamber has been one of the biggest opponents of EFCA. The group has lobbied heavily against the bill, arguing it would lead to job losses from strikes and union disruption. The labor movement considered the bill one of its top legislative priorities this Congress. Unions say it will help the economy by granting more workers collective bargaining rights, pushing up wages and benefits in the process. Despite the union support, EFCA has struggled to gain enough backing to move past an expected Republican-led filibuster in the Senate. Several Democratic centrists have distanced themselves from the bill, with some, such as Sen. Blanche Lincoln (D-Ark.), coming out against it. So far, EFCA has not had a vote in this Congress. Republicans and business groups, however, remain on alert because Democrats could take up the bill in Congress’s lame-duck session after the elections. Union leaders have hinted at that being a possibility.  \dots (\textbf{center})
        }& 
        Ours&
        \small 
        The Chamber has been one of the biggest opponents of EFCA.
        The group has argued the bill would lead to job losses from strikes and union disruption.
        Unions say it will help the economy by granting more workers collective bargaining rights.       
        \dots &
        center \cmark\\
         &
         \textsc{falcon}&
         \small Can you elaborate on the Chamber's argument that EFCA would lead to job losses from strikes and union disruption? How does the labor movement respond to this claim? The Chamber argues that EFCA would allow unions to quickly organize workers
         \dots & \textcolor{left}{left}
          \xmark\\
         &
         \textsc{llama2}&   
         \small The Chamber of Commerce has been actively campaigning against Democratic Senate candidates due to their support of the Employee Free Choice Act (EFCA). The Chamber believes the bill would lead to job losses and union disruption, while labor unions argue it would help the economy by granting more workers collective bargaining rights and \dots &
         center \cmark\\
         \bottomrule[1.5pt]
    \end{tabularx}
    }
    \caption{Example \#5 of one news article, three summaries generated by \ourmethod{} and two baselines, as well as their stances as evaluated by the political bias classifier.}
\end{table*}

\begin{table*}[t]
    \vspace{-60pt}
    \setlength{\tabcolsep}{15pt}
    \centering
    \resizebox{0.9\linewidth}{!}{
    \begin{tabularx}{\linewidth}{@{} p{6cm}| p{0.3cm}<{\centering} p{5cm}|m{0.1cm}<{\centering}}
        \toprule[1.5pt]
         Context&
         Model&
         Summary&
         Stance \\ \midrule[0.75pt]
         \vspace{-15pt}
         \multirow{3}{6cm}{
         \small Nancy Pelosi, Chuck Schumer demand equal airtime for Democrats after Donald Trump address. Democratic leaders on Monday asked the broadcast networks for equal airtime to counter President Trump’s planned immigration speech Tuesday, saying it’s only fair they get a chance to counter his “malice and misinformation.” House Speaker Nancy Pelosi and Senate Minority Leader Charles E. Schumer signaled disappointment with the networks’ decisions to grant the president a slot at 9 p.m.  \dots Democrats must immediately be given equal airtime,” they said in a joint statement. The minority party is usually granted time after major addressed such as a state of the union speech, though reactions to short presidential addresses to the nation are usually less structured. Mr. Trump in his speech is expected to make a plea for Congress to approve \dots Democrats have resisted, saying they won’t approve any new money for a program they say is unneeded and which Mrs. Pelosi calls “immoral.” \dots (\textbf{center})
        }& 
        Ours&
        \small 
        Democratic leaders asked the broadcast networks for equal airtime.
        Nancy Pelosi and Charles E. Schumer signaled disappointed with the networks' decisions to grant the president a slot \dots
        Obama again blamed him for the partial government shutdown \dots
        Democrats have resisted, saying they won’t approve any money for a program that is unneeded and which Mrs. Pelosi calls ‘immoral’ &
        \textcolor{left}{left} \xmark\\
         &
         \textsc{falcon}&
         \small Given that the President is making false claims about border ‘security’,” Mr. Schumer and Mrs. Pelosi said, “we cannot allow the President to use the airwaves, at a time of his choosing, to further mislead the American people.”
         \dots & \textcolor{left}{left}
          \xmark\\
         &
         \textsc{llama2}&   
         \small The president’s speech is expected to be carried live on all major television networks, including ABC, CBS, NBC, CNN and Fox News. The speech is scheduled to begin at 9 p.m. EST. \dots &
          \xmark\\
         \bottomrule[1.5pt]
    \end{tabularx}
    }
    \caption{Example \#6 of one news article, three summaries generated by \ourmethod{} and two baselines, as well as their stances as evaluated by the political bias classifier.}
\end{table*}

\end{document}